# Interactive and Scale Invariant Segmentation of the Rectum/Sigmoid via User-Defined Templates

Tobias Lüddemann[a] and Jan Egger[b,c]

[a] Technical University of Munich, Dept. of Mechatronics, Boltzmannstraße 15, 85748 Garching, Germany
[b] TU Graz, Institute for Computer Graphics and Vision, Inffeldgasse 16, 8010 Graz, Austria
[c] BioTechMed, Krenngasse 37/1, 8010 Graz, Austria

## ABSTRACT

Among all types of cancer, gynecological malignancies belong to the 4th most frequent type of cancer among women. Besides chemotherapy and external beam radiation, brachytherapy is the standard procedure for the treatment of these malignancies. In the progress of treatment planning, localization of the tumor as the target volume and adjacent organs of risks by segmentation is crucial to accomplish an optimal radiation distribution to the tumor while simultaneously preserving healthy tissue. Segmentation is performed manually and represents a time-consuming task in clinical daily routine. This study focuses on the segmentation of the rectum/sigmoid colon as an Organ-At-Risk in gynecological brachytherapy. The proposed segmentation method uses an interactive, graph-based segmentation scheme with a user-defined template. The scheme creates a directed two dimensional graph, followed by the minimal cost closed set computation on the graph, resulting in an outlining of the rectum. The graphs outline is dynamically adapted to the last calculated cut. Evaluation was performed by comparing manual segmentations of the rectum/sigmoid colon to results achieved with the proposed method. The comparison of the algorithmic to manual results yielded to a Dice Similarity Coefficient value of 83.85±4.08%, in comparison to 83.97±8.08% for the comparison of two manual segmentations of the same physician. Utilizing the proposed methodology resulted in a median time of 128 seconds per dataset, compared to 300 seconds needed for pure manual segmentation.

## 1. DESCRIPTION OF PURPOSE

Gynecological malignancies which include endometrial, vaginal/vulvar and cervical cancers represent the 4th most frequent type of cancer among women and a major cause of death around the world[1]. The standard procedure for primary or recurrent treatments of these types of cancer consists of external-beam radiation (EBR) followed by brachytherapy. During the brachytherapy procedure needle like catheters carrying a radiation source are inserted into the patient in close proximity to the tumor in order to directly irradiate the malignant tissue[2]. A crucial step in planning the applied amount and distribution of radiation is the segmentation of the tumor and adjacent organs-at-risk (OAR) potentially exposed to radiation. The most common OAR segmented in gynecological brachytherapy include the urinary bladder and the rectum/sigmoid colon. In general cases, the performing physician might have to outline several structures in more than 80 slices, which is a tedious task.

Over the last decades various fully automatic segmentation methods have been proposed to support this time-consuming segmentation process. However, the development of fully automatic segmentation tools remains problematic due to variability in pelvic organ shape and poor soft tissue depiction. None of these automatic analysis tools can guarantee robust results or achieved clinical approval. Moreover, they mostly lack necessary intervention methodologies and provide little final control by the medical doctor over the segmentation[3-7]. As none of the current fully automatic approaches provides sufficient results, manual refinement of the computed segmentation is desirable and the motivation for the development of interactive segmentation approaches[8-11]. In summary, they can speed up the segmentation process yet give the physician enough control over the algorithm to directly influence and improve the segmentation result.

This study focuses on the segmentation of the rectum/sigmoid colon as an organ-at-risk in the context of MR images for Image Guided Gynecological Brachytherapy. The rectum/sigmoid colon has – due to its mostly inhomogeneous appearance – proven to be challenging for current segmentation tools[12]. We address this problem by presenting an interactive and scale invariant segmentation algorithm for the longitudinal segmentation of the rectum/sigmoid colon based on graph theory. In addition, the approach uses a user-defined template for the graph construction to handle the variations in anatomy which is to the best of our knowledge a novelty in literature.

---

E-mail: jan.egger@icg.tugraz.at

## 2. METHODS

The methodology is based on a graph network and has been developed during a German diploma thesis[13]; preliminary results have been presented in a congress abstract[14] and at a German conference[15]. The approach is a consistent further development of previous publications where different templates have been used to create a graph[16-25]. However, in these publications fixed pre-defined shapes have been used for the segmentation process, e.g. a square template for 2D vertebra segmentation in sagittal slices. Thus, these approaches were not able to handle segmentations of structures that vary in anatomy shape from patient to patient. In summary, we solve this problem by letting the user define an individual template by simply outlining the structure contour in the first slice (Figure 1). Thereafter, this initial information is used to automatically construct a specific graph to segment consecutive slices (Figure 2). In order to define the graph's shape, diameter and area to spread within, a graph template $T \in R^3$ is introduced: $T = \{M_0,...,M_n\} : 0 \leq n \leq \infty$. The markers $M_t = (m_x, m_y, m_z)$ of template $T$ are vectors in $R^3$, where $\forall m_z := \phi$, resulting in a two dimensional template. The markers $M_t$ of template $T$ are denoted by $M_t = (T)$. A given template $T$, either user drawn or the last calculated min-cut[26], will be scaled by a specific size $s$ in order to be used as template for a next min-cut. Next, the nodes $V$ of the graph $G(V,E)$ are sampled according to the template. From a given used-defined seed point $SP = \{sp_x, sp_y, sp_z\}$ a number of rays $r_i : 1 \leq i \leq k$ are spread in uniform angles. For each ray $r_i$ the intersection point $IP$ with $T$ is calculated. The length of each ray $r_i$ is defined by the distance between $SP$ and the intersection point $IP$ of $T$ : $|r_i| = |IP_{ri} - SP|$. Along all rays $r_i$ a number of nodes $V_{i,n} : 0 \leq n \leq u$ is distributed in equal distances from the seed point to the template border. Moreover, the graph network requires the implementation of two distinct virtual nodes that are not represented by a voxel of image $I$, the source $s$ and the sink $t$. So far the nodes $V$ of the graph $G(V,E)$ have been implemented. The next step to set up the complete graph network is to implement the graph's edges $E$. In the following, three different types of edges can be distinguished (see also the notation of Li et al.[27]): **Z-weighted edges** are intra-edges connecting the seed point $SP$ with the first node $v_{i,0}$ of each ray $r_i$. They also connect all neighboring nodes $(v_{i,n}, v_{i+1,n})$ of the same ray $r_i$. They are denoted by the subset $E_z \subset E$; **XY-weighted edges** are inter-edges connecting nodes $(v_{i,n}, v_{i+1,m})$ of directly neighboring rays $r_i, r_{i+1}$. They are denoted by the subset $E_{xy} \subset E$; **T-weighted edges** connect each node $v_{i,n} \in V(G)$ with the source $s$ and the sink $t$. They are denoted by the subset $E_T \subset E$. Next, weights are assigned to the different edges: to ensure a minimal closed set, the capacity $c(v_{i,n}, v_{i+1,n})$ for all edges $(v_{i,n}, v_{i+1,n}) \in E_z$ (**Z-weights**) is set to a maximum value indicated by $\infty$. The capacity value assigned to each **T-weight** is determined by the intensity contrast of the grey value $gv(v_{i,n})$ of the currently regarded node $v_{i,n}$ to the previous node $v_{i,n-1}$ on the same ray. The **XY-weights** $E_{XY} \in E(G)$ can be considered as stiffness parameters depending on the parameter $\Delta$, which realizes different smoothness values for the resulting minimal cut and greater variability among possible *cuts* with the same cost amount. However, rugged templates showed to be disadvantageous for the segmentation result, such that the use of parameter $\Delta$ was limited to values $\Delta \leq 2$. The calculated *cut* of a specific slice is stored as segmentation result together with all previous cuts of this object. After the complete object is segmented, slices that have been skipped by the user have to be interpolated in order to generate an object contour in each image slice. Thereafter, the set of object contours is used for voxelization of the object and to generate a 3D Object. The set of object contours is stored as cso file, while the voxelized 3D object is stored as nrrd file for further use in treatment planning.

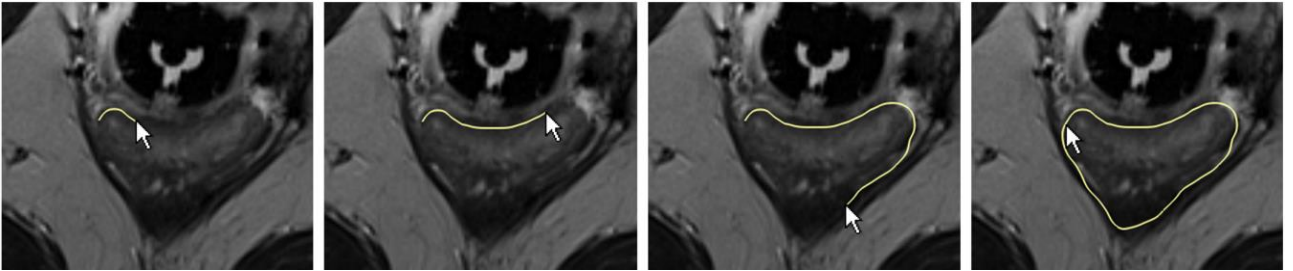

Fig. 1: The rectum is manually outlined (yellow) to define a user-defined template for a segmentation graph.

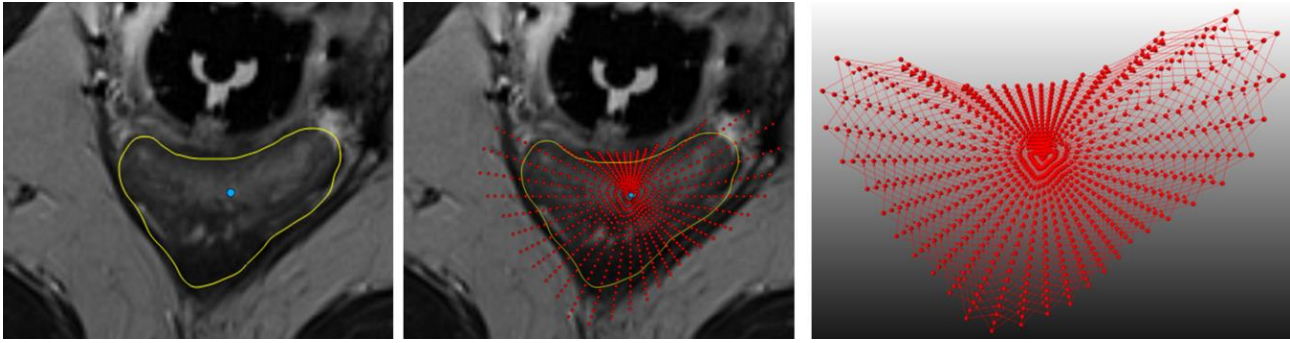

Fig. 2: Graph construction via a user-defined template: the manually outlined rectum (yellow) is used to calculate the graph's center point (blue, left image), next the graph's nodes are sampled (red, middle image) and finally the whole graph is constructed (red, right image)

## 3. RESULTS

For evaluation of the introduced approach, datasets have been used where the appearance of the rectum/sigmoid colon strongly varies. From each dataset to another, its shape, size, slice thickness and grey value distribution alter in no predictable manner. The data used for this experiment is a set of seven MRI datasets acquired during gynecological interstitial brachytherapy cases[28-30]. For comparison of the algorithmic computed segmentation result, a medical doctor experienced in radiological segmentations of gynecological brachytherapy manually outlined the rectum/sigmoid two times with a time difference of several months. Computation was performed in MeVisLab[31] on a personal computer with an Intel® Core™ i3 CPU M330 with 2.13 GHz dual core, which enabled the interactive algorithm to run smoothly without any delay or interruptions. The interactive segmentation was performed with a constant *t*-weight parameter of 0.2. The scaling factor *sf* and the number of rays *r* and points per ray *n* were $sf \epsilon [1.6]$, $r \epsilon \{40\}$ and $n \epsilon \{40\}$. In order to estimate the complexity of the segmentation task, the respective number of slices where the segmentation was performed as well as the object's volume (cm³) and number of voxels were investigated. Furthermore, the time in seconds for the segmentation of each dataset was acquired by screen capture recordings. According to the performing physician, the segmentation per dataset took in average 300 sec (5 min). In comparison to the manual expert segmentation, the presented algorithm segmented a smaller volume in all cases except case 7, where a slightly larger volume was segmented. Furthermore, the algorithm achieved a median segmentation time per dataset of 128 seconds compared to the manual segmentation, which represents a ~60% time saving. The semi-automatic segmentation results for the datasets 1-7 have been compared to the expert manual segmentation results M2. The algorithm achieved an average DSC of 83.85±4.08% when compared to M2 (Table 1). The two manual expert segmentations M1 and M2 have been compared in an intra-analysis and a DSC has been calculated in order to precisely asses the DSC value of the interactive result. The intra-analysis between manual segmentation dataset M1 and M2 yielded a DSC of 83.97±8.08%. This value reflects the difficulty inherent in the segmentation of the rectum/sigmoid colon.

| Data set | DSC (%) | | Hausdorff Dist. (Voxel) | |
|---|---|---|---|---|
| | IC – M2 | M1 – M2 | IC – M2 | M1 – M2 |
| 1 | 88.43 | 86.93 | 11.04 | 4.03 |
| 2 | 80.88 | 85.16 | 6.48 | 11.45 |
| 3 | 79.04 | 78.19 | 25.47 | 18.92 |
| 4 | 80.17 | 70.37 | 11.05 | 22.29 |
| 5 | 84.78 | n.a. | 9.34 | n.a. |
| 6 | 89.54 | 91.05 | 4.36 | 9.78 |
| 7 | 84.14 | 91.40 | 9.64 | 4.12 |
| μ±σ | 83.85±4.08 | 83.97±8.08 | 11.05±6.81 | 11.76±7.54 |
| min | 79.04 | 70.37 | 4.36 | 4.03 |
| max | 89.54 | 91.40 | 25.47 | 22.29 |
| Table 1: Comparison of the interactive results with expert manual segmentation results. IC indicates the interactive result, M the respective manual expert segmentation, n.a. indicates non availability of the respective dataset. | | | | |

Figure 3 and Figure 4 show segmentation results taken from dataset 1 and 3. The red area represents the segmentation result of the interactive segmentation, whereas the green area represents the manual expert result. The image (3) in Figure 3 depicts the voxelized segmentation result of a manual segmentation and image (3) in Figure 4 depicts an interactive segmentation result.

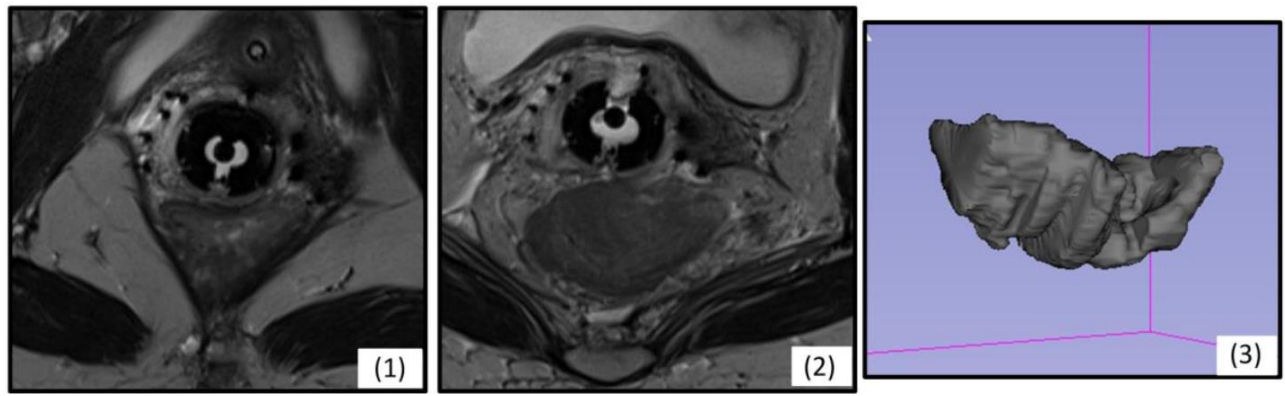

Fig. 3: Exemplary axial slices for dataset one (1) and three (2). On the right side (3) the voxelized masks of a manual segmentation is shown.

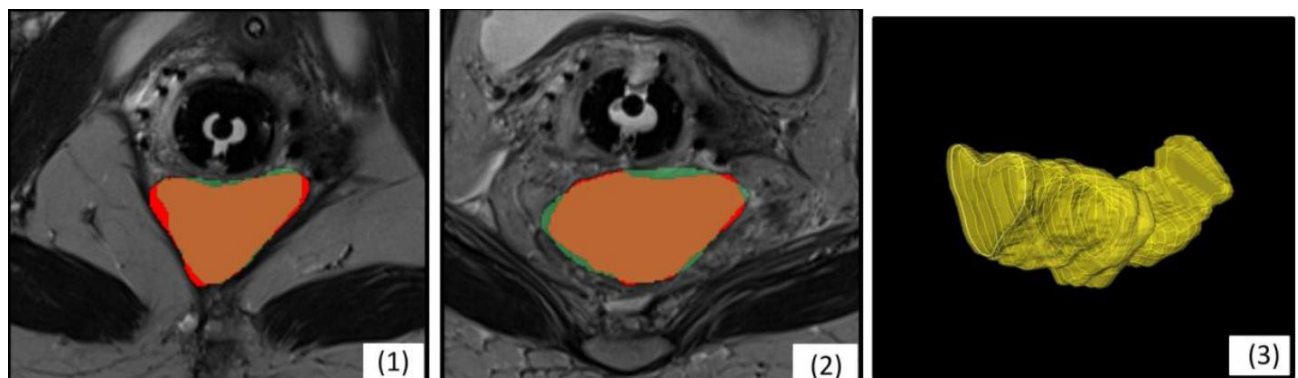

Fig. 4: Segmentation results for dataset one (1) and three (2). Red shows the interactive and green the manual segmentations, and brown the overlap. On the right side the voxelized masks of an interactive (3) segmentation is shown.

## 4. CONCLUSIONS

The objective of this contribution was the development of a novel interactive graph-based segmentation methodology for the rectum/sigmoid colon as an Organ-At-Risk in the context of interstitial gynecological brachytherapy. The clinical gold standard for segmentation tasks is given by time and resource consuming slice-by-slice manual outlining of the region of interest. Although various fully automatic segmentation tools have been developed up to date, none of these have found their way into clinical practice. One drawback of automatic approaches is their complexity, limited flexibility and little provided user influence on the computed segmentation result where the algorithm is challenged. Thus, manual refinement of the computed segmentation is desirable and the motivation for the development of interactive segmentation approaches that allow direct and intuitive control over the segmentation result. Evaluation of the algorithm was performed by comparing the computer assisted segmentation results with manual expert segmentations of the rectum/sigmoid colon and yielded to a DSC of 83.85±4.08%. An intra-analysis between two sets of manual expert segmentations resulted to a DSC of 83.97±8.08. The average time for the segmentation of a dataset dropped from approximately 300 to 128 seconds for the rectum/sigmoid colon utilizing the presented methodology.

There are several areas of future work: due to the star-like graph node distribution with the seed point at its origin, the algorithm was challenged where the object shape was particularly concave or convex. For future approaches multiple, equally distributed seed points could be utilized to guarantee an ideal areal sampling of the object and thus enhance the segmentation quality. Another idea is to enhance the approach to 3D by outlining the structure in an axial, sagittal and coronal slice and creating a 3D graph for an interactive segmentation with this information. Finally, the approach can be used to segment other longitudinal/tubular structures, like fiber tracts[32-34] or the aorta to support the time-consuming analysis of aortic aneurysms[35-39], which then can be used for virtual stenting[40-43].

## ACKNOWLEDGEMENT

DDr. Jan Egger receives funding from BioTechMed-Graz ("Hardware accelerated intelligent medical imaging"). A video demonstrating the iterative segmentation can be found under the following YouTube-channel:

https://www.youtube.com/c/JanEgger/